\documentclass[letterpaper]{article} 
\usepackage{aaai25}  
\usepackage{times}  
\usepackage{helvet}  
\usepackage{courier}  
\usepackage[hyphens]{url}  
\usepackage{graphicx} 
\usepackage{natbib}  
\usepackage{caption} 
\usepackage{algorithm}
\usepackage{algorithmic}
\usepackage{amsmath}
\usepackage{booktabs}
\usepackage{multirow}
\usepackage{pifont}
\usepackage{amsmath}
\usepackage{graphicx}
\usepackage{microtype}
\usepackage{amssymb}
\newcommand{\cmark}{\ding{51}}
\newcommand{\xmark}{\ding{55}}
\usepackage{listings}
\usepackage{enumitem}
\usepackage{xcolor}
\usepackage{soul}
\usepackage{newfloat}
\usepackage{listings}
\DeclareCaptionStyle{ruled}{labelfont=normalfont,labelsep=colon,strut=off} 
\floatstyle{ruled}
\newfloat{listing}{tb}{lst}{}
\floatname{listing}{Listing}
\title{TCPFormer: Learning Temporal Correlation with Implicit Pose Proxy for \\ 3D Human Pose Estimation}
\author{
    Jiajie Liu\textsuperscript{\rm 1}, Mengyuan Liu\textsuperscript{\rm 1}\thanks{Corresponding author: liumengyuan@pku.edu.cn}, Hong Liu\textsuperscript{\rm 1}, Wenhao Li\textsuperscript{\rm 2}
}
\affiliations{
    \textsuperscript{\rm 1}State Key Laboratory of General Artificial Intelligence, Peking University, Shenzhen Graduate School\\
    \textsuperscript{\rm 2}Nanyang Technological University\\

}

\usepackage{bibentry}

\begin{document}

\maketitle

\begin{abstract}
Recent multi-frame lifting methods have dominated the 3D human pose estimation.
However, previous methods ignore the intricate dependence within the 2D pose sequence and learn single temporal correlation.
To alleviate this limitation, we propose TCPFormer, which leverages an implicit pose proxy as an intermediate representation.
Each proxy within the implicit pose proxy can build one temporal correlation therefore helping us learn more comprehensive temporal correlation of human motion.
Specifically, our method consists of three key components: Proxy Update Module (PUM),  Proxy Invocation Module (PIM), and Proxy Attention Module (PAM). 
PUM first uses pose features to update the implicit pose proxy, enabling it to store representative information from the pose sequence.
PIM then invocates and integrates the pose proxy with the pose sequence to enhance the motion semantics of each pose.
Finally, PAM leverages the above mapping between the pose sequence and pose proxy to enhance the temporal correlation of the whole pose sequence.
Experiments on the Human3.6M and MPI-INF-3DHP datasets demonstrate that our proposed TCPFormer outperforms the previous state-of-the-art methods. 
\end{abstract}

%
\begin{links}
    \link{Code}{https://github.com/AsukaCamellia/TCPFormer}
\end{links}

\section{Introduction}

3D human pose estimation has always been a crucial problem in computer vision, which aims to locate the 3D joint positions of a human body~\cite{ moon2020i2l, pavlakos2018ordinal, chen2021anatomy}. Nowadays, 3D human pose estimation finds widespread applications in various scenarios, including motion prediction~\cite{wang2023dynamic}, action recognition~\cite{zhang2022unsupervised}, and human-robot interaction~\cite{ gong2022meta, ye2021collaborative}. Given the widespread usage of 2D human pose detectors~\cite{chen2018cascaded, he2017mask, newell2016stacked, sun2019deep} and the task-relatedness between 2D pose and 3D pose, most research follows a 2D-to-3D lifting pipeline~\cite{poseformer,stride,mhformer,mixste,wang2024skeleton}, where 2D keypoints are first detected and then lifted to the 3D space. Despite the considerable success achieved, this task remains an ill-posed problem and inherently suffers from depth ambiguity.
 \begin{figure}[t] \centering
\includegraphics[width=1\linewidth]{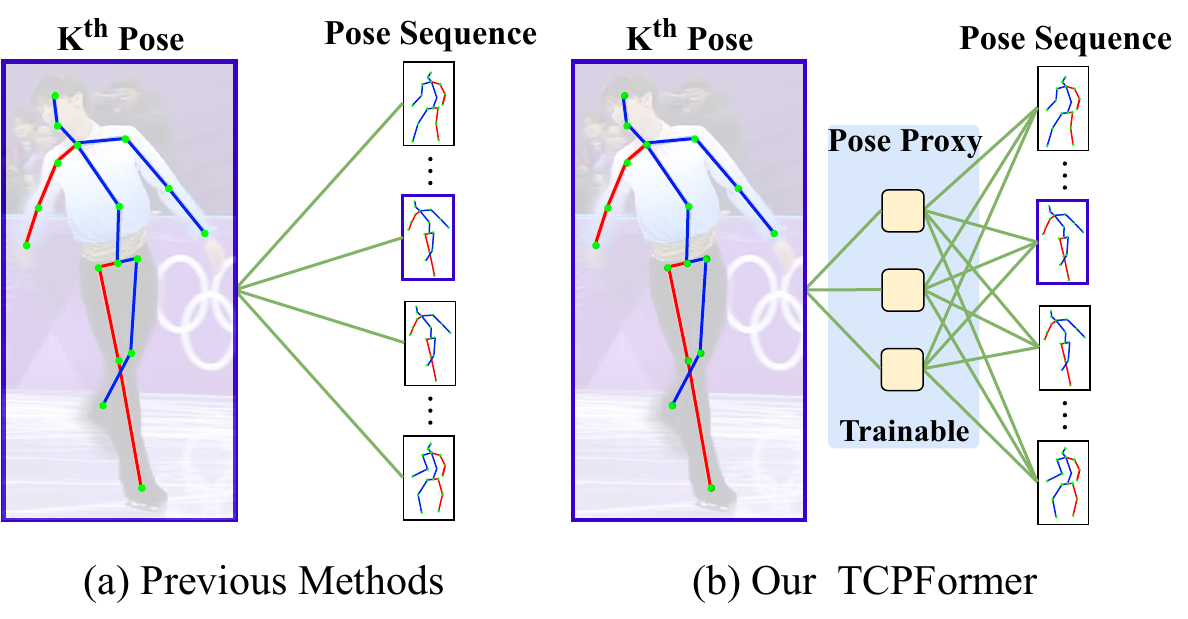}
\caption{\textbf{An illustration of our motivation.} Given a pose sequence of length \textit{T}, we take the individual pose within the pose sequence as an example.
(a) In previous methods, one pose establishes the temporal correlation with the pose sequence only in one \textit{1-to-T} mapping.
(b) We introduce an implicit pose proxy to act as an intermediate representation.
Each proxy within the implicit pose proxy of length \textit{L} can establish one \textit{1-to-T} mapping, which facilitates learning more comprehensive temporal correlation.
}
\label{fig:1}
\end{figure}
An extensive body of literature focuses on exploiting temporal information between adjacent frames to mitigate this issue, ranging from earlier methods~\cite{videopose, liu2020attention, chen2021anatomy} use temporal convolution and subsequent attempts~\cite{cai2019exploiting, hu2021conditional, wang2020motion} use graph convolution.
Recently, transformer~\cite{transformer} have achieved significant success in both natural language preprocess~\cite{ brown2020language, devlin2018bert} and computer vision~\cite{ ViT, carion2020end}. For the 3D human pose estimation task, many works~\cite{mhformer,mixste,pstmo,tang20233d} leverage the powerful sequence modeling capability of transformer to extend their input from the limited neighboring frames to long-term sequences for advanced accuracy, e.g., 243 video frames for MixSTE~\cite{mixste} and STCFormer~\cite{tang20233d}; large as 351 frames for MHFormer~\cite{mhformer}.

Despite their achievements, a potential concern has gradually emerged: with the massive increase in the number of input frames, the performance improvement becomes slow. For instance, PoseFormerV2~\cite{poseformerv2} achieved an error reduction of 0.8mm when expanding the input from 81 frames to 243 frames. StridedTrans~\cite{stride} achieved a marginal 0.3mm error reduction when expanding the input from 243 frames to 351 frames, while MHFormer~\cite{mhformer} achieved an even smaller error reduction of 0.2mm with the same input expansion. These key observations point towards a problem that restricts most methods from effectively modeling the temporal correlation within the 2D pose sequence. In this work, we are trying to solve this problem.

As illustrated in Figure~\ref{fig:1}a, we discover that most of the aforementioned multi-frame methods only establish one \textit{1-to-T} mapping for each pose within the pose sequence, where \textit{T} denotes the length of pose sequence. 
However, due to the extensive number of frames, only establishing one \textit{1-to-T} mapping can not comprehensively reflect the complex temporal correspondence within the pose sequence.

To address this limitation, we propose a novel method to learn \textbf{T}emporal \textbf{C}orrelation with Implicit Pose \textbf{P}roxy, dubbed \textbf{TCPFormer}. As illustrated in Figure~\ref{fig:1}b, we introduce an implicit pose proxy to act as the intermediate representation. 
We first establish a \textit{1-to-L} mapping to build the relationship between the individual pose and the implicit pose proxy, where \textit{L} denotes the length of the implicit pose proxy. Then, each proxy within the implicit pose proxy will interact with the pose sequence and build multiple \textit{1-to-T} mapping to help model learning more comprehensive temporal correlation.
Moreover, our implicit pose proxy is trainable and will be continuously optimized during the training process.

Specifically, we first propose a Proxy Update Module (PUM). PUM adaptively encodes useful and representative information from the pose sequence to update the pose proxy through the cross-attention mechanism \cite{transformer}. Although the information in the pose proxy has been updated, we have not yet transmitted it to each pose within the pose sequence.
Therefore, we propose a Proxy Invocation Module (PIM) that uses the pose proxy as the key and value to enhance the feature representation ability of the pose sequence.
In addition, we propose a Proxy Attention Module (PAM). PAM skillfully leverages the two cross-attention matrices of PUM and PIM to get an aggregation matrix and flexibly fuses it with the original self-attention matrix to obtain a more effective and comprehensive temporal correlation.

We extensively evaluate our TCPFormer on two widely used benchmark datasets, Human3.6M~\cite{h36m} and MPI-INF-3DHP~\cite{3dhp}.
Empirical evaluations show that our approach outperforms the previous state-of-the-art methods. 
Comprehensive ablation studies are also presented to evaluate the contribution of each component. Our \textbf{contributions} can be summarized as follows:
\begin{itemize}
\item To the best of our knowledge, we are the first to introduce the implicit pose proxy to 3D human pose estimation.
Our method leverages the implicit pose proxy as an intermediate representation to effectively model the complex temporal correlation within the pose sequence.
\item We design three novel modules: Proxy Update Module, Proxy Invocation Module, and Proxy Attention Module. These three modules present a unique way to effectively enhance the feature of pose sequence and learn a more comprehensive temporal correlation.
\item Extensive experiments conducted on Human3.6M and MPI-INF-3DHP two challenging datasets for 3D human pose estimation demonstrate
that our method achieves superior performances than the previous methods.
\end{itemize}

\section{Related Work}

\subsection{3D Human Pose Estimation} 
Early works~\cite{ ionescu2014iterated, h36m, ramakrishna2012reconstructing, agarwal2005recovering, andriluka2009pictorial, ionescu2011latent} of monocular 3D human pose estimation primarily focus on exploiting spatial prior information in the form of human skeletal structure and motion features. With the development of deep learning, more deep neural network-based methods have been introduced and can be divided into two mainstream types: one-stage manner and two-stage manner. One-stages approaches~\cite {kanazawa2018end, pavlakos2017coarse, sun2018integral} directly estimate the 3D pose from the input image without the intermediate 2D pose representation.
Different from the one-stage manner, two-stage methods~\cite{fang2018learning, simplebaseline, zhao2019semantic, liu2020comprehensive, xu2021graph} first obtain 2D joint coordinates in the image and then leverage the task-relevant positional information to lift the 2D joint coordinates to 3D poses. 
With the reliable achievement of 2D human pose detectors~\cite{chen2018cascaded, he2017mask, newell2016stacked, sun2019deep}, these 2D-to-3D lifting methods outperform one-stage approaches. However, they still inherently suffer from the problem of depth ambiguities.
To address this problem, some studies~\cite{liu2020attention,videopose} have made preliminary explorations in utilizing temporal information. Liu \textit{et al.}~\cite{ liu2020attention} extend the temporal convolutional network by introducing the attention mechanism. 
The aforementioned methods utilize limited temporal information, which is unable to effectively facilitate 3D human pose estimation.

\subsection{Transformer-based Methods} 
For the 3D human pose estimation task, PoseFormer~\cite{ poseformer} firstly introduces transformer architecture to leverage spatial and temporal dependency. 
MHFormer~\cite{mhformer} addresses the depth ambiguity by learning spatio-temporal representations of multiple pose hypotheses.
MixSTE~\cite{mixste} constructs a mixed spatio-temporal transformer to capture the temporal motion of different body joints.
P-STMO~\cite{pstmo} is the first approach that introduces the pre-training technique to 3D human pose estimation.
PoseFormerV2~\cite{poseformerv2} improves PoseFormer by utilizing a frequency-domain representation of input joint sequences.
STCFormer~\cite{tang20233d} decomposes spatio-temporal attention and integrates the structure-enhanced positional embedding.
MotionBERT~\cite{motionbert} and UPS~\cite{foo2023unified} both train a unified model for multi-task.
However, these methods still have limitation in directly modeling the complex temporal correlation of the pose sequence due to sequence length. We conduct further exploration to address this limitation in this paper.

\label{sec:overview}
\begin{figure*}[t] \centering
\includegraphics[width=1\linewidth]{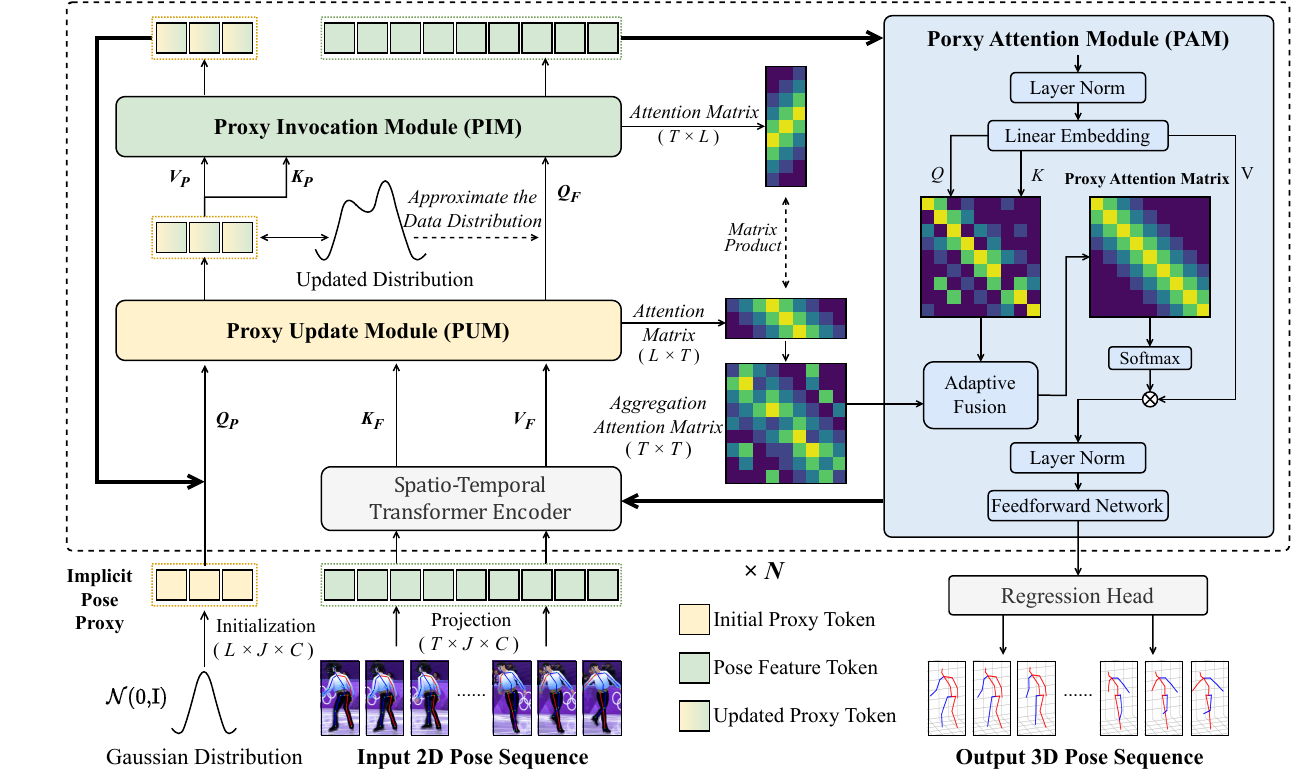}
\caption{
\textbf{Overview of our method.}
We first extract the spatio-temporal information through a spatio-temporal encoder. 
Then, we introduce an implicit pose proxy which is initialized by Gaussian distribution.
These features and proxy are then handed to the proxy update module to update the implicit pose proxy. Next, the proxy invocation module uses the updated pose proxy to enhance the feature of the pose sequence. We obtain an aggregation attention matrix through two cross attention matrices and send it with the pose sequence feature to the proxy attention module to learn comprehensive temporal correlation.
After repeating the above processes \textit{N} times, we use a regression head to obtain the 3D pose sequence.
}
\label{fig:overview}
\end{figure*}

\section{Method}

Given a 2D pose sequence $X \in \mathbb{R}^{T \times J \times C_{\text{in}}}$, our goal is to estimate the 3D pose sequence $Y \in \mathbb{R}^{T \times J \times C_{\text{out}}}$. Here, $T$ refers to the number of input frames, and $J$ refers to the number of joints. $ C_{\text{in}}$ and $C_{\text{out}}$ denote the dimension of input 2D pose and output 3D pose. 

\subsection{Overview}
An overview of our pipeline is illustrated in Fig.~\ref{fig:overview}. Firstly, we use a base spatio-temporal encoder to extract information from $X$ and obtain a basic feature $B\in\mathbb{R}^{T \times J \times C_{f}}$, where $C_f$ denotes the dimension of the hidden feature. 
We reshape $B$ into $F\in\mathbb{R}^{J \times T \times C_f}$ as the per-joint temporal features. $F$ is then fed into our Proxy Update Module with the implicit pose proxy $P\in\mathbb{R}^{J \times L \times C_f}$ which is initialized by Gaussian distribution. Here, $L$ indicates the time dimension of implicit pose proxy (further discussion in the ablation study). This step will update implicit pose proxy $P$.
Then, the updated implicit pose proxy $P$ and feature $F$ will be sent to our Proxy Invocation Module to generate the enhanced features $\widetilde{F}\in \mathbb{R}^{J \times T \times C_f}$. 
During above process, we can obtain the cross-attention matrix $M_{\text{$P$$\rightarrow$$F$}}\in \mathbb{R}^{L \times T }$ and $M_{\text{$F$$\rightarrow$$P$}}\in\mathbb{R}^{T \times L}$. 
We aggregate them to generate a aggregation attention matrix $M_{\text{$F$$\rightarrow$$F$}}\in\mathbb{R}^{T \times T}$.
The enhanced feature $\widetilde{F}$ and the aggregation attention matrix  $M_{\text{$F$$\rightarrow$$F$}}\in\mathbb{R}^{T \times T}$ are sent to Proxy Attention Module to produce the final feature $\overline{F}\in \mathbb{R}^{J \times T \times C_f}$. In the end, we use a regression head to generate the final 3D pose sequence $Y$.

\subsection{Proxy Update Module}
\label{sec:mrmi}

For the proxy update module, we first use three linear matrices $W^{U}_{Q}$,$W^{U}_{K}$ and $W^{U}_{V}$ to map proxy $P$ to queries $Q_{P}$, the input feature $F$ to keys $K_{F}$ and values $V_{F}$ as follows:
\begin{equation}
Q_{P}=P\cdot W^{U}_{Q} \quad   K_{F}=F\cdot W^{U}_{K} \quad V_{F}=F\cdot W^{U}_{V}
\end{equation}
Then we use the Softmax attenion~\cite{transformer}, which has been widely used in modern transformer designs to get the cross attention matrix $M_{\text{$P$$\rightarrow$$F$}}$ as follows:
\begin{equation}
M_{P \rightarrow F} = Softmax({Q_{P} \cdot K^{T}_{F}}/{\sqrt{C_{f}}}) \label{eq2}
\end{equation}
Finally, we fuse the $M_{\text{$P$$\rightarrow$$F$}}$ and  $V_{F}$ to get the updated implicit pose proxy $P$ as follows:
\begin{equation}
P = P +M_{P \rightarrow F}\cdot V_{F} \label{eq3}
\end{equation}

\subsection{Proxy Invocation Module}
After updating the implicit pose proxy, We use inverse processes to integrate the updated implicit pose proxy $P$ into the feature $F$. Similarly, we use three linear matrices $W^{I}_{Q}$,$W^{I}_{K}$ and $W^{I}_{V}$ to project the feature $F$ to queries $Q_{F}$, the updated implicit pose proxy $P$ to keys $K_{P}$, and values $V_{P}$ as:
\begin{equation}
Q_{F}=F\cdot W^{I}_{Q} \quad   K_{P}=P\cdot W^{I}_{K} \quad V_{P}=P\cdot W^{I}_{V}
\end{equation}
Then, we use the Softmax function to get the cross attention matrix $M_{\text{$F$$\rightarrow$$P$}}$ as follows:
\begin{equation}
M_{F \rightarrow P} = Softmax({Q_{F} \cdot K^{T}_{P}}/{\sqrt{C_{f}}}) 
\end{equation}
Finally, we utlize the $M_{\text{$F$$\rightarrow$$P$}}$ and  $V_{P}$ to get the enhanced pose sequence feature $\widetilde{F}$ as follows:
\begin{equation}
\widetilde{F} = F +M_{F \rightarrow P}\cdot V_{P} 
\end{equation}

\subsection{Proxy Attention Module}
\label{sec:ma}

After the above process, we can obtain two cross-attention matrices $M_{F \rightarrow P}$ and $M_{P \rightarrow F}$. We skillfully leverage them to further learn temporal correlation.
Concretely, we use matrix multiplication to obtain an aggregation attention matrix as:
\begin{equation}
\label{eq8}
M = M_{F \rightarrow P} \cdot M_{P \rightarrow F}
\end{equation}
Then, we map the enhanced feature $\widetilde{F}$ to queries $Q_{\widetilde{F}}$, keys $K_{\widetilde{F}}$, and values $V_{\widetilde{F}}$ as follows: 
\begin{equation}
Q_{\widetilde{F}}=\widetilde{F}\cdot W^{A}_{Q} \quad   K_{\widetilde{F}}=\widetilde{F}\cdot W^{A}_{K} \quad V_{\widetilde{F}}=\widetilde{F}\cdot W^{A}_{V}
\end{equation}
The aggregation attention matrix is then adaptively fused with the original self attention matrix as follows:
\begin{equation}
    \overline{M} = Sigmoid(\mu)\cdot M + (1 -Sigmoid(\mu) ) \cdot Q_{\widetilde{F}}\cdot K^{T}_{\widetilde{F}}
\end{equation}
where $\mu$ is a learnable parameter for each layer during the training process to adaptively learn suitable fusion ratios.
Finally, we get the final pose sequence feature $\overline{F}$ as follows:
\begin{equation}
\overline{F} = \widetilde{F} + Softmax(\overline{M}/\sqrt{C_{f}}) \cdot V_{\widetilde{F}} 
\end{equation}


\subsection{Regression Head and Loss Function}
\label{sec:loss}
After the above process is repeated several times, we project feature $\overline{F}$ to a higher dimension by applying a linear layer and tanh activation to compute the motion semantics and use a linear transformation layer as regression head to estimate the final 3D pose sequence $Y$.

Our network has two optimization objectives and is trained in an end-to-end manner. We use L2 loss to minimize the errors between predictions and ground truths:
\begin{equation}
 \mathcal{L}_{3D}=\frac{1}{JT} \sum_{j=1}^{J} \sum_{t=1}^{T} \left\|\widehat{Y}_{j,t}-Y_{j,t}\right\|_{2}
\end{equation}
where $\widehat{Y}_{j,t}$ and $Y_{j,t}$ are the ground truth and estimated 3D pose of the $j$-th joint in $t$-th frame.
In addition, the temporal consistency loss (TCLoss) in~\cite{hossain2018exploiting} is introduced to produce smooth poses. Specifically, the TCLoss can be formulated as follows:
\begin{equation}
\mathcal{L}_{T}=\frac{1}{J(T-1)}  \sum_{j=1}^{J} \sum_{t=2}^{T}    \left\|   \Delta \widehat{Y}_{j,t}  -\Delta  Y_{j,t}\right\|_{2}
\end{equation}
where $\Delta\widehat{Y}_{t} = \widehat{Y}_{t} - \widehat{Y}_{t-1}$, $\Delta Y_{t} = Y_{t} - Y_{t-1}$.
The final loss function $\mathcal{L}$ is then defined as :
\begin{equation}
\mathcal{L} = \mathcal{L}_{3D}+\lambda \mathcal{L}_{T}
\end{equation}
where $\lambda$ is used to balance position accuracy and motion smoothness.

\section{Experiments}

\label{sec:blind}

\subsection{Datasets and Evaluation Metrics}
We comprehensively evaluate our model on two large-scale 3D human pose estimation datasets: Human3.6M~\cite{h36m} and MPI-INF-3DHP~\cite{3dhp}. 

\noindent\textbf{Human3.6M} is the most popular benchmark for indoor 3D human pose estimation, which contains approximately 3.6 million frames captured by 4 cameras at different views. This dataset contains 11 subjects performing 15 typical actions.

\noindent\textbf{MPI-INF-3DHP} is a recently proposed large-scale challenging dataset with both indoor and outdoor scenes. The training set comprises 8 subjects, covering 8 activities, ranging from walking and sitting to complex exercise poses and dynamic actions. The test set covers 7 activities, containing three scenes: green screen, non-green screen, and outdoor environments.

\noindent\textbf{Evaluation Metrics.} For the Human3.6M dataset, we use two evaluation metrics: MPJPE and P-MPJPE. MPJPE (Mean Per Joint Position Error) is computed as the mean Euclidean distance between the estimated joints and the ground truth in millimeters after aligning their root joints. P-MPJPE (Procrustes-MPJPE) is the MPJPE after the estimated joints align to the ground truth via a rigid transformation.
For the MPI-INF-3DHP dataset, following previous works~\cite{pstmo,zhou2024lifting,motionbert}, we use ground
truth 2D pose as input and report MPJPE, Percentage of
Correct Keypoint (PCK) with the threshold of 150mm, and Area Under Curve (AUC) as the evaluation metrics.

\subsection{Implementation Details}
We consider the layers $N$ of modules, the number $H$ of heads in attention block, the size $C$ of hidden feature, the temporal dimension $L$ of implicit pose proxy, and the initialization distribution $D$ of proxy as free parameters. The performances of the versions with ($N = 16, H = 8, C = 128, L = T/3, D = Gaussian$) are reported. Our model is implemented using PyTorch and executed on a server equipped with 2 NVIDIA 4090 GPUs. We apply horizontal flipping augmentation for both training and testing following~\cite{tang20233d,motionbert,foo2023unified,zhao2023contextaware}. For model training, we set each mini-batch as 16 sequences. The network parameters are optimized using AdamW~\cite{adamw} optimizer over 90 epochs with a weight decay of 0.01. The initial learning rate is set to 5e-4 with an exponential learning rate decay schedule and the decay factor is 0.99. In the experiments on Human3.6M, two kinds of input are utilized including the 2D ground truth and the Stacked Hourglass~\cite{newell2016stacked} 2D pose detection, following~\cite{motionbert,ci2019optimizing}. For MPI-INF-3DHP, 2D ground truth is used following previous works~\cite{cai2024disentangled,mixste,li2023pose,motionbert,hot}.

\begin{table*}[t]
\setlength{\tabcolsep}{1.5mm} 

  \centering
  \fontsize{9}{10}\selectfont{
  \begin{tabular}{l|c|cc|ccccc}
    \toprule
    Method   &  Venue   & Seq2Seq& $T$ &Parameter&MACs &MACs/frames &MPJPE $\downarrow$&P-MPJPE $\downarrow$ \\
    \midrule

MHFormer~\cite{mhformer}& CVPR'22& \xmark&351& 30.9M&7.1G &7096M &43.0&34.4\\
MixSTE~\cite{mixste} &CVPR'22& \cmark&243& 33.6M& 139.0G&572M & 40.9&32.6\\
P-STMO~\cite{pstmo} &ECCV'22& \xmark&243& 6.2M&0.7G &740M & 42.8&34.4\\

STCFormer~\cite{tang20233d}& CVPR'23& \cmark&243& 4.7M&19.6G & 80M& 41.0&32.0\\
PoseFormerV2~\cite{poseformerv2}& CVPR'23& \xmark&243& 14.3M&0.5G & 528M& 45.2&35.6\\
GLA-GCN~\cite{yu2023gla} &ICCV'23& \xmark&243&1.3M &1.5G &1556M & 44.4&34.8\\
MotionBERT~\cite{motionbert} &ICCV'23& \cmark&243&42.3M & 174.8G& 719M& \underline{39.2}&32.9\\
KTPFormer~\cite{peng2024ktpformer} &CVPR'24& \cmark&243&33.7M & 69.5G& 286M& 40.1&\underline{31.9}\\

\midrule
\textbf{TCPFormer (Ours)} &-& \cmark&81&35.0M & 36.4G& 150M& 40.5&33.7\\
\textbf{TCPFormer (Ours)} &-& \cmark&243&35.1M & 109.2G& 449M& \textbf{37.9}&\textbf{31.7}\\
    \bottomrule
  \end{tabular}
  }
  \caption{Quantitative comparisons on Human3.6M dataset. $T$ is the number of input frames. Seq2seq refers to estimating 3D pose sequences rather than only the center frame. MACs/frames represents multiply-accumulate operations for each output frame. The best result is shown in bold, and the second-best result is underlined. Our TCPFormer achieves the best performance with smaller parameters and computational cost compared with MotionBERT~\cite{motionbert}. 
  }
  \label{tab:param}
\end{table*}

\begin{table*}[t]
\begin{center}
\setlength{\tabcolsep}{0.5mm} 

\fontsize{9}{10}\selectfont{
\begin{tabular}{l|c|ccccccccccccccc|c}
\toprule

\textbf{MPJPE (GT)} &$T$ & Dir. & Disc. & Eat & Greet & Phone & Photo & Pose & Pur. & Sit & SitD. & Smoke & Wait & WalkD. & Walk & WalkT. & Avg \\
\midrule

MHFormer~\cite{mhformer}&351 &27.7 & 32.1 & 29.1 & 28.9 & 30.0 & 33.9 & 33.0 & 31.2 & 37.0 & 39.3 & 30.0 & 31.0 & 29.4 & 22.2 & 23.0&30.5 \\

MixSTE~\cite{mixste}&243 &21.6 & 22.0 & 20.4 & 21.0 & 20.8 & 24.3 & 24.7 & 21.9 & 26.9 & 24.9 & 21.2 & 21.5 & 20.8 & 14.7 & 15.7 & 21.6 \\
P-STMO~\cite{pstmo} & 243 &28.5 & 30.1 & 28.6 & 27.9 & 29.8 & 33.2 & 31.3 & 27.8 & 36.0 & 37.4 & 29.7 & 29.5 & 28.1 & 21.0 & 21.0 & 29.3 \\

STCFormer~\cite{tang20233d} &243 &20.8 & 21.8 & 20.0 & 20.6 & 23.4 & 25.0 & 23.6 & 19.3 & 27.8 & 26.1 & 21.6 & 20.6 & 19.5 & 14.3 & 15.1 & 21.3\\
PoseFormerV2~\cite{tang20233d}& 243 &-&-&-&-&-&-&-&-&-&-&-&-&-&-&-&- \\

GLA-GCN~\cite{yu2023gla} & 243 &20.1 & 21.2 & 20.0 & 19.6 & 21.5 & 26.7 & 23.3 & 19.8 & 27.0 & 29.4 & 20.8 & 20.1 & 19.2 & 12.8 & 13.8 & 21.0 \\

MotionBERT~\cite{motionbert} &243 &\underline{16.7} & 19.9 & \underline{17.1} & \underline{16.5} & \underline{17.4} & \underline{18.8} & \underline{19.3} & 20.5 & 24.0 & \underline{22.1} & \underline{18.6} & \underline{16.8} & \underline{16.7} & \underline{10.8} & \underline{11.5} & \underline{17.8} \\
~\cite{peng2024ktpformer} &243&19.6&\underline{18.6}&18.5&18.1&18.7&22.1&20.8&\underline{18.3}&\underline{22.8}&22.4&18.8&18.1&18.4&13.9&15.2&19.0\\
\midrule
\textbf{TCPFormer (Ours)} & 81 &19.2&19.6&20.2&18.3&20.7&24.0&20.9&20.3&26.9&26.9&21.6&18.5&19.6&15.7&15.4&20.5 \\
\textbf{TCPFormer (Ours)} &243  &\textbf{15.0}&\textbf{15.9}&\textbf{16.1}&\textbf{14.2}&\textbf{15.7}&\textbf{16.4}&\textbf{16.2}&\textbf{16.9}&\textbf{22.1}&\textbf{20.6}&\textbf{16.7}&\textbf{13.3}&\textbf{13.8}&\textbf{9.2}&\textbf{9.9}&\textbf{15.5}\\

\bottomrule
\end{tabular}
}
\caption{Results on Human3.6M dataset in millimeters under MPJPE using ground truth 2D pose as input. $T$ is the number of input frames. The best result is shown in bold, and the second-best result is underlined. Our TCPFormer reduces 2.3mm (12.9\%) MPJPE compared with MotionBERT~\cite{motionbert} and achieves the best performance in all actions.}
        \label{tab:h36mres}
\end{center}

\end{table*}

\begin{table*}[h]
  \centering

\label{tab:3dhp}
\setlength{\tabcolsep}{3.500mm} 
\fontsize{10}{11}\selectfont{
    \begin{tabular}{l|c|cc|ccc}
    \toprule
 Method&Venue&$T$ &Seq2Seq&PCK $\uparrow$&AUC $\uparrow$&MPJPE $\downarrow$\\
\midrule
    MHFormer~\cite{mhformer}& CVPR'22     & 9 &\xmark & 93.8&63.3&58.0\\
    MixSTE~\cite{mixste}& CVPR'22     & 27 &\cmark & 94.4&66.5&54.9\\
    P-STMO~\cite{pstmo}& ECCV'22     & 81 &\xmark & 97.9&75.8&32.2\\
    STCFormer~\cite{tang20233d}& CVPR'23     & 81 &\cmark & 98.7&83.9&23.1\\
    PoseFormerV2~\cite{poseformerv2} &CVPR'23     & 81&\xmark  & 97.9&78.8&27.8\\
    GLA-GCN~\cite{yu2023gla} &ICCV'23     & 81 &\xmark & 98.5&79.1&27.8\\
MotionBERT~\cite{motionbert} &ICCV'23     & - &\cmark & -&-&-\\

    KTPFormer~\cite{peng2024ktpformer}& CVPR'24     & 81 &\cmark & \underline{98.9}&85.9&\underline{16.7}\\
    \midrule
    \textbf{TCPFormer (Ours)} & -& 9&\cmark&98.3&84.4&20.4\\
    \textbf{TCPFormer (Ours)} & -& 27&\cmark&98.7&\underline{86.5}&17.8\\
    \textbf{TCPFormer (Ours)} & -& 81&\cmark&\textbf{99.0}&\textbf{87.7}&\textbf{15.0}\\
\bottomrule
    \end{tabular}
    }
\caption{Results on MPI-INF-3DHP under three evaluation metrics. $T$ is the number of input frames. Seq2seq refers to estimating 3D pose sequences rather than only the center frame. The best result is shown in bold, and the second-best result is underlined.}
\label{tab:3dhp}
\end{table*}

\subsection{Comparison with State-of-the-art Methods}
Due to page limitations, we only present the best results of other methods along with their corresponding number of input frames (\textit{T}). Our TCPFormer achieves the best performance across different numbers of input frames.

\noindent\textbf{Results on Human3.6M.} We compare our TCPFormer with previous methods on the Human3.6M dataset. 
Table~\ref{tab:param} summarizes the performance comparisons in terms of MPJPE and P-MPJPE errors of all 15 actions and the computational cost of each method. Our method achieves state-of-the-art performance with an MPJPE of 37.9mm and P-MPJPE of 31.7mm with $T$ = 243. It is worth noting that our method with $T$ = 81 input frames still achieves competitive performance with an MPJPE error of 40.5mm and surpasses the performance of most methods with a higher number of input frames.
For example, this result outperforms PoseformerV2~\cite{poseformerv2} (40.5mm v.s. 45.2mm) with 243 frames, and MHFormer~\cite{mhformer} even with 351 frames (40.5mm v.s. 43.0mm). Our model also achieves a good balance between performance and computational cost. TCPFormer achieves the best performance with lower parameters (35.1M v.s. 42.5M) and MACs per frame (449M v.s. 719M) compared with MotionBERT~\cite{motionbert}.

To explore the lower bound of our method, we also directly use the 2D ground truth as input. As shown in the Table~\ref{tab:h36mres}, our method with $T$ = 243 achieves the best performance with an MPJPE of 15.5mm in all actions.
Similarly, our method with $T$ = 81 input frames surpasses the performance of most methods with a higher number of input frames. For example, this result outperforms STCFormer~\cite{tang20233d} (20.5mm v.s. 21.3mm) with 243 frames, and MHFormer~\cite{mhformer} with 351 frames (20.5mm v.s. 30.5mm).

\noindent\textbf{Results on MPI-INF-3DHP.} To demonstrate the generalization capability of our model, we evaluate our model on the challenging MPI-INF-3DHP dataset, which includes more complex scenes and motions. Following previous works~\cite{poseformer,mixste,pstmo,hot}, we use ground truth 2D as input and set the number of input frames as 9, 27, or 81 due to the shorter video sequences. As observed in Table~\ref{tab:3dhp}, our method with $T$ = 81 achieves the best performance with the PCK of 99.0\%, AUC of 87.7\%, and  MPJPE of 15.0mm. More remarkably, our method with $T$ = 9 input frames still outperforms the previous state-of-the-art model STCFormer~\cite{tang20233d} with $T$ = 81 input frames, despite having only one-ninth of the input frames (9 frames v.s. 81 frames).

\subsection{Ablation Study}
\label{sec:ablation}

All experiments were conducted on the Human3.6M dataset with $T$ = 243 as the number of input frames.

\begin{table}[h]
  \centering
\setlength{\tabcolsep}{1.25mm} 
  \fontsize{10}{11}\selectfont{
\begin{tabular}{c|cccc|cc}
    \toprule
    Step&Proxy&PUM&PIM&PAM&MPJPE $\downarrow$&P-MPJPE $\downarrow$\\
    \midrule
    1&\cmark&-&-&-&42.2&34.6\\
    2&\cmark&\cmark&-&-&39.5&32.6\\
    3&\cmark&\cmark&\cmark&-&38.7&32.3\\
    Ours&\cmark&\cmark&\cmark&\cmark&\textbf{37.9}&\textbf{31.7}\\

    \bottomrule
\end{tabular}
}
  \caption{The effectiveness of different components. All our proposed novel components exhibit improvements. }
  \label{tab:component}
\end{table}

\noindent\textbf{Impact of Each Component.} As shown in Table~\ref{tab:component}, we validate the overall performance gain brought by the proposed implicit pose proxy (Proxy), proxy update module (PUM), proxy invocation module (PIM), and proxy attention module (PAM). 
Our baseline, which only introduces an implicit pose proxy without additional module design, achieves a result of 42.2mm MPJPE and 34.6mm P-MPJPE.
By applying PUM, our method decreases 2.7mm MPJPE and 2.0mm P-MPJPE.
Next, we integrate PIM into our method and achieve better results with 38.7mm MPJPE and 32.3mm P-MPJPE.
Finally, we achieve the best performance with 37.9mm MPJPE and 31.7mm P-MPJPE by incorporating the PAM.

\begin{table}[h]
\centering
\setlength{\tabcolsep}{3.00mm} 
  \fontsize{10}{11}\selectfont{\begin{tabular}{cc|cc}
\toprule
Length&Distribution&MPJPE $\downarrow$&P-MPJPE $\downarrow$\\
\midrule
27&Gaussian&38.4&32.1\\
81&Random&39.1&32.5\\
81&Laplacian&38.2&32.0\\
\textbf{81}&\textbf{Gaussian}&\textbf{37.9}&\textbf{31.7}\\
243&Gaussian&38.6&32.7\\
\bottomrule
\end{tabular}}
  \caption{Analysis on implicit pose proxy. Distribution and Length denote the temporal dimension and initial distribution of our proposed implicit pose proxy.
  }
  \label{tab:proxy}
\end{table}

\noindent\textbf{Analysis on Implicit Pose Proxy.} How to represent implicit pose proxy is crucial for our methods. We investigated the impact of the temporal dimension and initial distribution of our implicit pose proxy. 
For the temporal dimension, we set it to 27, 81, and 243 respectively. 
For the initial distribution, we provided gaussian distribution, laplace distribution, and random distribution.
The results presented in Table~\ref{tab:proxy} show that our method achieves the best performance when setting the temporal dimension of implicit pose proxy to 81 and using the gaussian distribution initialization.

\noindent\textbf{Analysis on Micro Design.}  In this section, we further explore the effectiveness of various micro designs within
the proxy update module (PUM) and proxy invocation module (PIM). As shown in Table~\ref{tab:pumpim}, we achieve the best performance when both PUM and PIM use cross attention.

\begin{table}[t]
\centering
\setlength{\tabcolsep}{1.4mm} 
  \fontsize{9}{10}\selectfont{\begin{tabular}{c|c|cc}
\toprule
PIM &PUM&MPJPE $\downarrow$&P-MPJPE $\downarrow$ \\
\midrule
 MLP &MLP & 40.8&33.8\\
 CrossAttention &MLP & 39.6&32.6\\
 MLP &CrossAttention & 39.5&32.8\\

\textbf{CrossAttention} &\textbf{CrossAttention} & \textbf{38.7}&\textbf{32.3}\\
\bottomrule
\end{tabular}}
\caption{Analysis of the various micro designs within proxy update module and proxy invocation module. 
}
\label{tab:pumpim}

\end{table}

\begin{table}[t]
\centering
\setlength{\tabcolsep}{3.00mm} 
  \fontsize{10}{11}\selectfont{\begin{tabular}{cc|cc}
\toprule
Range &Strategy&MPJPE $\downarrow$&P-MPJPE $\downarrow$\\
\midrule
0&Fixed&38.4&32.4\\
0&Trainable&38.2&32.0\\
\textbf{(0, 1)}&\textbf{Trainable}&\textbf{37.9}&\textbf{31.7}\\
(-1, 0)&Trainable&38.5&32.3\\
(-1, 1)&Trainable&37.9&32.0\\
\bottomrule
\end{tabular}}
  \caption{Analysis on the adaptive fusion. Range denotes the sampling range of $\mu$. Strategy denotes whether $\mu$ is trainable. 
  }
  \label{tab:mam}
\end{table}

\noindent\textbf{Analysis on Proxy Attention Module.}
We extensively investigated the fusion strategies of adaptive fusion within our proxy attention module.
Specifically, we pay attention to the sampling range of $\mu$ and whether it is trainable. As shown in Table~\ref{tab:mam}, we achieved the best performance when we allowed $\mu$ to be trainable and sampled it from (0, 1).

\begin{figure*}[h] \centering
\includegraphics[width=1\linewidth]{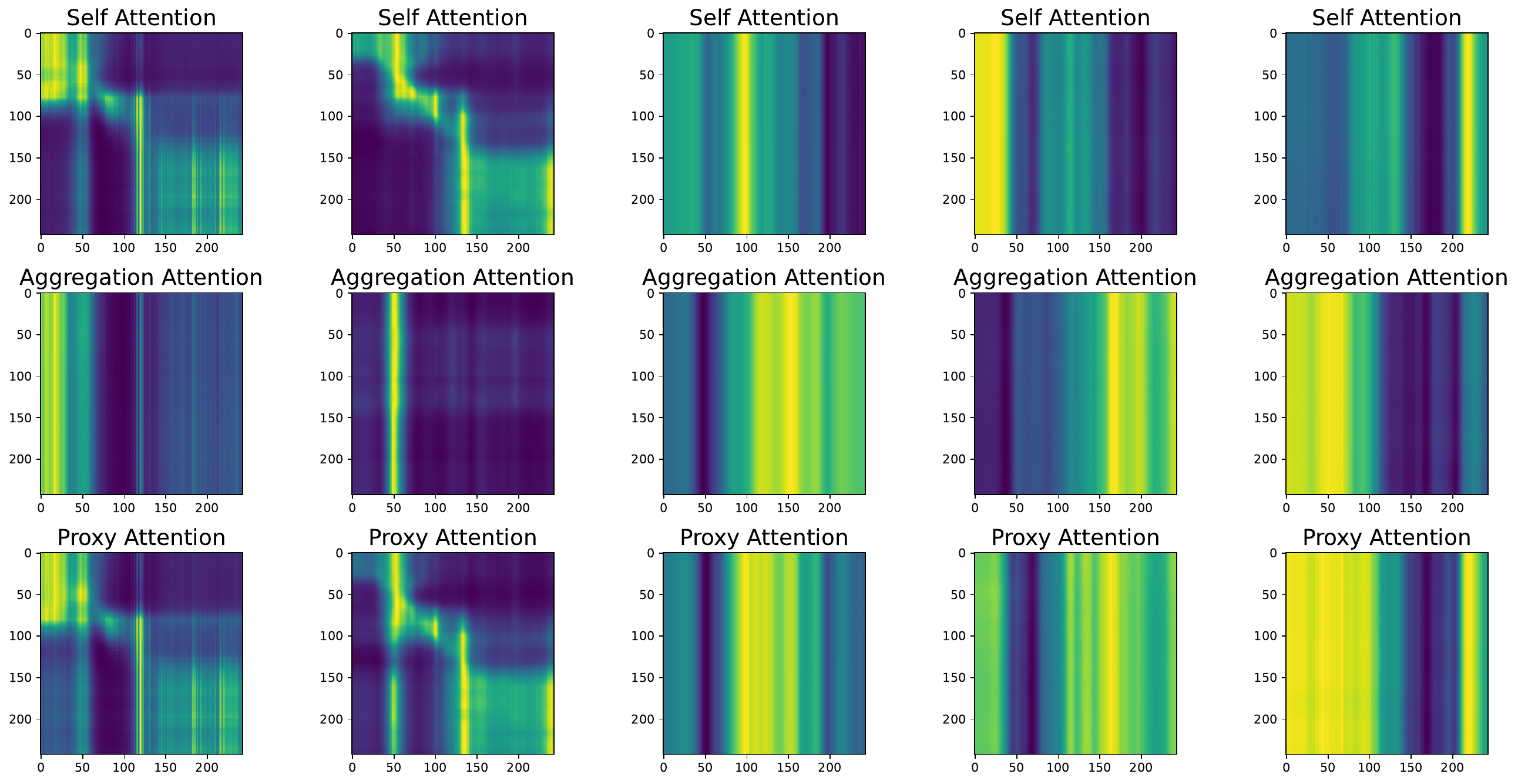}
\caption{
Visualizations of different attention matrices. The first row is the original self-attention matrix. The second row is the aggregation attention matrix. The third row is our proxy attention matrix. As expected, our proxy attention matrix effectively leverages the aggregation attention matrix to complement the missing parts of the original self attention matrix.
}
\label{fig:vis}
\end{figure*}

\begin{figure}[t] \centering
\includegraphics[width=1\linewidth]{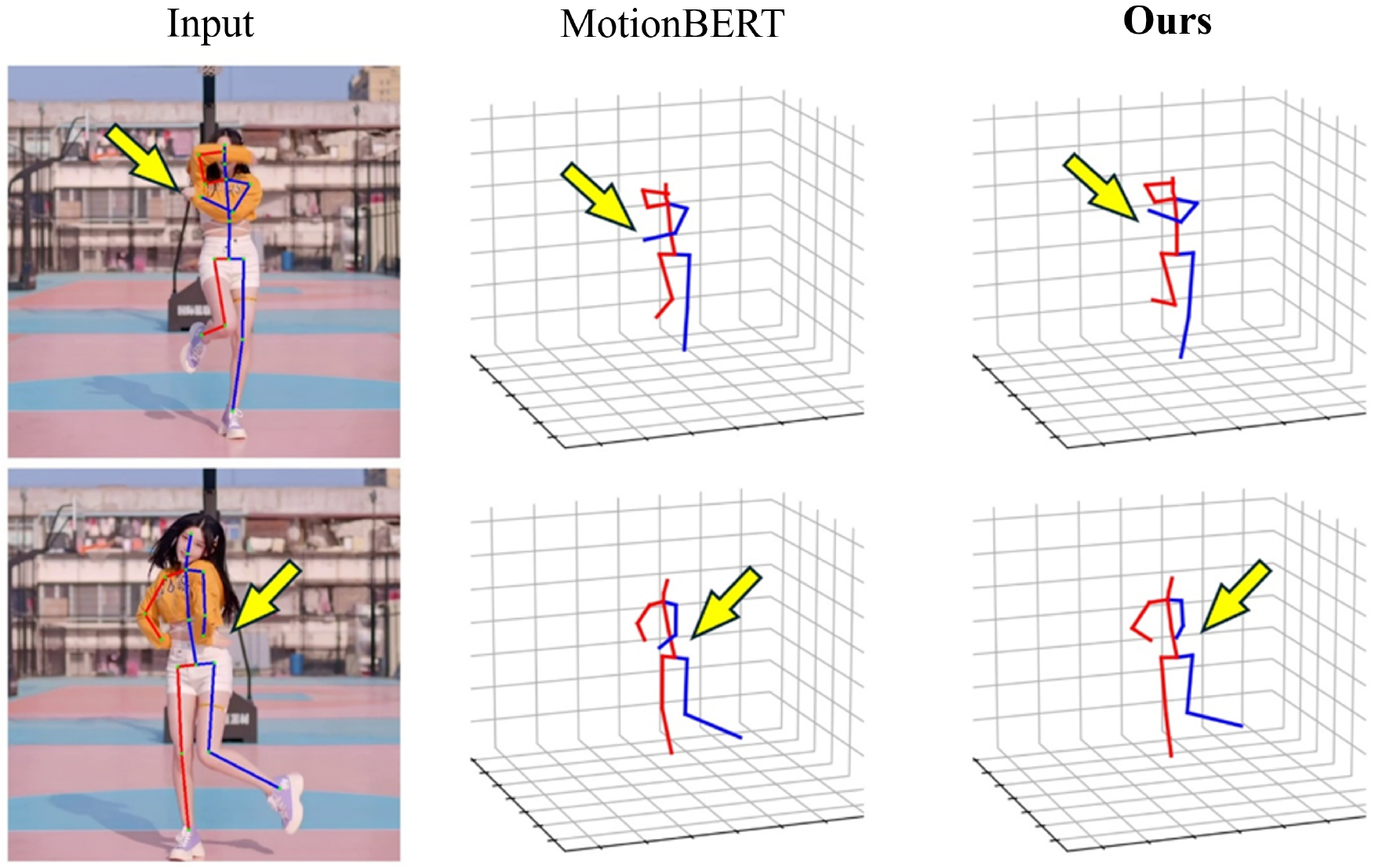}
\caption{Qualitative comparisons of our TCPFormer with MotionBERT on in-the-wild videos. The yellow arrows indicate locations where our method achieves better results.
}
\label{fig:pose}
\end{figure}

\begin{figure}[t] \centering
\includegraphics[width=1\linewidth]{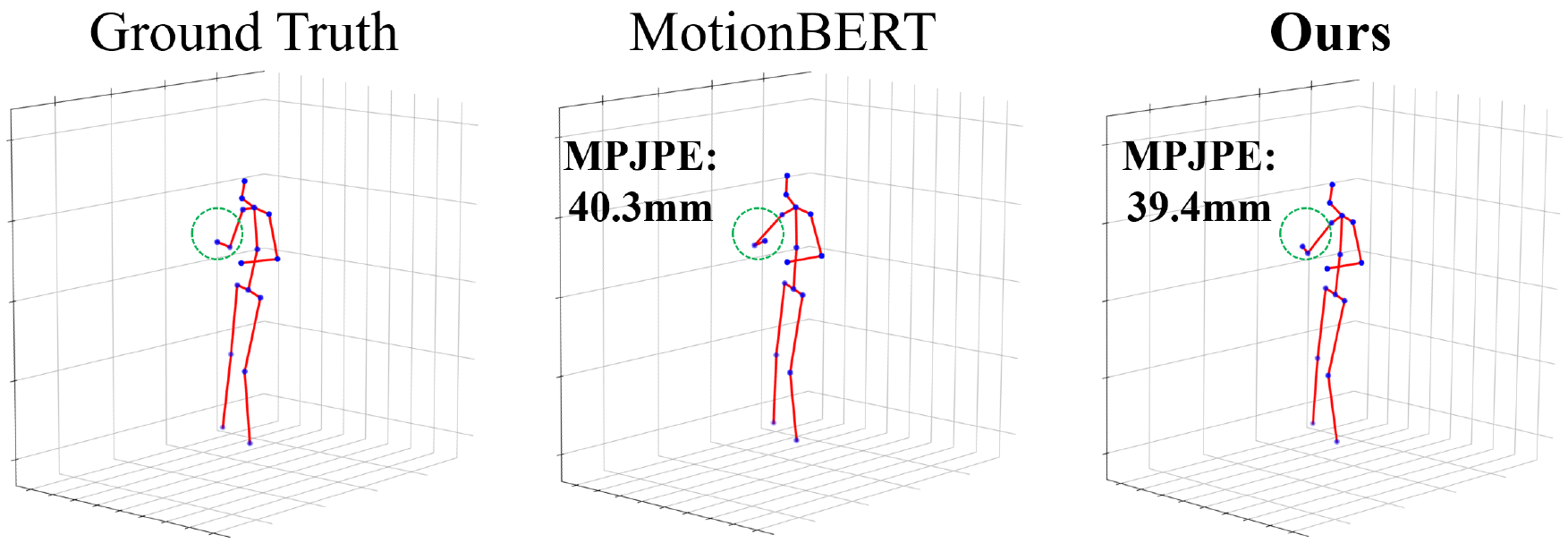}
\caption{Qualitative comparisons of our TCPformer with MotionBERT on Human3.6M. The green circles indicate locations where our method achieves better results.
}
\label{fig:h36mpose}
\end{figure}

\noindent\textbf{Qualitative Analysis.}
We visualized the original self attention matrix (first row), aggregation attention matrix (second row), and proxy attention matrix (third row) in Figure~\ref{fig:vis}. All attention matrices are normalized to $[0,1]$. 
As expected, our proxy attention matrix effectively leverages the aggregation attention matrix to complement the original self attention matrix.
Furthermore, we also present 3D human pose estimation results by MotionBERT~\cite{motionbert} and our TCPFormer on the Human3.6M dataset and in-the-wild videos.  
As shown in Figure~\ref{fig:pose} and Figure~\ref{fig:h36mpose}, TCPFormer achieves better qualitative results compared with MotionBERT~\cite{motionbert}.

\section{Conclusion}
In this paper, we present TCPFormer, a novel method to learn temporal correlation with implicit pose proxy. 
Different from previous methods that learn complex temporal correlations only through single mapping, TCPFormer leverages the implicit pose proxy as an intermediate representation to skillfully model the complex temporal correlation within the pose sequence and effectively use the temporal information to facilitate 3D human pose estimation. 
The visualization results provide empirical evidence that our TCPFormer can build comprehensive temporal correlation within the 2D pose sequence.
Extensive experimental results also show that our TCPFormer outperforms the previous state-of-the-art approaches on the Human3.6M and MPI-INF-3DHP datasets. 

\section{Acknowledgments}
This work was supported by National Natural Science Foundation of China (No. 62203476), Natural Science Foundation of Guangdong Province (No. 2024A1515012089), Shenzhen Innovation in Science and Technology Foundation for The Excellent Youth Scholars (No. RCYX20231211090248064).

\bibliography{aaai25}

\end{document}